\documentclass[table]{article}
\usepackage{tikz}
\usetikzlibrary{shapes.geometric, arrows}

\tikzstyle{startstop} = [rectangle, rounded corners, minimum width=3cm, minimum height=1cm,text centered, draw=black, fill=white!30]
\tikzstyle{salto} = [circle, minimum width=1.5cm, minimum height=1.5cm, text centered, draw=black, fill=white!30] 
\tikzstyle{process} = [rectangle, minimum width=4cm, minimum height=1.25cm, text centered, text width = 3.75cm, draw=black]
\tikzstyle{decision} = [diamond, minimum width=3cm, minimum height=1cm, text centered, draw=black, fill=white!30]
\tikzstyle{arrow} = [thick,->,>=stealth]

\usepackage{arxiv}
\usepackage[utf8]{inputenc} 
\usepackage[T1]{fontenc}    
\usepackage{hyperref}       
\usepackage{url}            
\usepackage{booktabs}       
\usepackage{amsfonts}       
\usepackage{nicefrac}       
\usepackage{microtype}      
\usepackage{lipsum}		
\usepackage{graphicx}
\usepackage{natbib}
\usepackage{doi}

\usepackage[spanish,es-tabla]{babel}

\usepackage{xcolor}

\usepackage{wrapfig}
\usepackage{caption}
\usetikzlibrary{decorations.pathreplacing}

\title{Detección y Cuantificación de Erosión Fluvial con Visión Artificial}

\author{    
    Paúl Maji, Marlon Túquerres, Stalin Valencia, Marcela Valenzuela y Christian Mejía-Escobar\\
    \\
Facultad de Ingeniería en Geología, Minas, Petróleos y Ambiental (FIGEMPA)\\
Universidad Central del Ecuador\\
Quito, Ecuador \\
\texttt{\{pdmaji,matuquerres,spvalencia,mbvalenzuela,cimejia\}@uce.edu.ec} \\
}

\renewcommand{\shorttitle}{\textit{Detección y Cuantificación de Erosión Fluvial con Visión Artificial}}

\hypersetup{
pdftitle={A template for the arxiv style},
pdfsubject={q-bio.NC, q-bio.QM},
pdfauthor={David S.~Hippocampus, Elias D.~Striatum},
pdfkeywords={First keyword, Second keyword, More},
}

\begin{document}
\maketitle

\begin{abstract}
La erosión fluvial es un proceso natural que puede generar impactos significativos en la estabilidad del suelo y las infraestructuras estratégicas.
La detección y monitoreo de este fenómeno tradicionalmente se abordan mediante métodos fotogramétricos y análisis en sistemas de información geográfica.
Estas tareas demandan conocimientos específicos y procesamiento manual intensivo.
Este estudio propone un enfoque basado en inteligencia artificial para la identificación automática de zonas erosionadas y la estimación de su área.
Se emplea el modelo de vanguardia en visión artificial YOLOv11, ajustado mediante fine-tuning y entrenado con fotografías e imágenes LiDAR. Este dataset combinado fue segmentado y etiquetado utilizando la plataforma Roboflow.
Los resultados experimentales indican una detección eficiente de patrones de erosión con una precisión del 70\%, una identificación acertada de las áreas erosionadas y un cálculo confiable de su extensión en píxeles y metros cuadrados.
Como producto final, se ha desarrollado el sistema EROSCAN, una aplicación web interactiva que permite a los usuarios cargar imágenes y obtener segmentaciones automáticas de erosión fluvial, junto con el área estimada.
Esta herramienta optimiza la detección y cuantificación del fenómeno, facilitando la toma de decisiones en gestión de riesgos y planificación territorial.
\end{abstract}


\section{Introducción}
\label{sec:intro}
El poder invisible que tiene el agua para dar forma a los ríos y transformar paisajes es conocido como \textit{erosión fluvial} \citep{GK}. Este proceso geológico natural puede convertirse en una amenaza crítica para la estabilidad de los suelos, la seguridad humana e infraestructuras cercanas \citep{Meneses}. Por esta razón, la detección temprana del frente erosivo resulta esencial para reducir el riesgo de desastres y prevenir pérdidas tanto humanas como económicas \citep{Comision2}. Además, el monitoreo de estos fenómenos geológicos tiene aplicaciones clave en la planificación urbana, la gestión de riesgos naturales, la conservación ambiental y la protección de infraestructuras estratégicas como carreteras, represas y oleoductos \citep{Comision}\citep{Naranjo}.

Sin embargo, monitorear eficazmente estos procesos implica enfrentar importantes desafíos técnicos y metodológicos. El principal inconveniente son los cambios en la morfología fluvial, los cuales pueden ocurrir de forma súbita tras eventos de crecidas, de forma progresiva o por procesos geomorfológicos que alteren el equilibrio del cauce \citep{Barrera}. Esta variabilidad complica la identificación de patrones de comportamiento, dificultando la toma de decisiones oportunas \citep{Jativa}.
Los métodos tradicionales abordan esta problemática incluyendo la recopilación de datos de campo, revisión de imágenes satelitales, toma de datos fotogramétricos y LiDAR, y su análisis realizado normalmente en Sistemas de Información Geográfica \citep{Graw}.
No obstante, estos métodos presentan varias limitaciones; por ejemplo, el uso de un software especializado donde dicho procesamiento requiere de un experto que dedique excesivo tiempo y esfuerzo humano para estas labores de interpretación de imágenes. 

Esta investigación aborda tales limitaciones utilizando Inteligencia Artificial (IA), que se ha convertido en una alternativa exitosa para el tratamiento de problemas relacionados con la naturaleza.
La identificación automática de zonas de erosión fluvial a través de imágenes es un problema que puede ser enmarcado en el campo de la visión por computadora, específicamente, como una tarea de detección y segmentación de objetos.
Estas tareas constituyen la especialidad de herramientas de deep learning de vanguardia como YOLO \cite{yolov11}, que es uno de los modelos más exitosos y populares según el estado del arte.

La estrategia del presente trabajo consiste en reutilizar este modelo, el cual originalmente fue entrenado con millones de imágenes genéricas y de múltiples categorías, y afinarlo con un conjunto de imágenes obtenidas a partir de fotogrametría y LiDAR específicas del fenómeno de erosión fluvial para alcanzar el objetivo de identificar las zonas erosionadas y estimar el área de afectación.

Este enfoque permite optimizar el análisis de la erosión, facilitando la identificación de patrones de cambios morfológicos y zonas de mayor riesgo de forma más rápida, precisa y con menos intervención de especialistas técnicos. Por tanto, es una herramienta útil para la toma de decisiones preventivas y reducción de los costos de mitigación.


Las contribuciones concretas de este estudio son: (1) una solución automatizada que integra un modelo de aprendizaje profundo para visión artificial con un módulo de cálculo del área afectada; (2) un conjunto de datos específico, depurado y etiquetado, para un entrenamiento orientado a erosión fluvial; y (3) una aplicación web desarrollada en Streamlit, que permite a los usuarios interactuar con el modelo de forma sencilla y eficaz.

El resto del manuscrito se organiza de la siguiente manera: la Sección \ref{sec:sota} explora los trabajos relacionados con nuestra problemática; la Sección \ref{sec:method} describe la creación del conjunto de datos y la generación del modelo; la Sección \ref{sec:experimental} presenta la experimentación realizada y los resultados obtenidos; la Sección \ref{sec:discussion} discute los hallazgos del estudio; la Sección \ref{sec:eroscan} expone el desarrollo de la aplicación web. Finalmente, la Sección \ref{sec:conclusion} enuncia las conclusiones, recomendaciones y líneas de trabajo futuro.

\section{Trabajos relacionados}
\label{sec:sota}

Nuestra búsqueda en la literatura científica no ha encontrado investigaciones específicas sobre la detección automática de erosión fluvial.
Sin embargo, existen varios estudios que emplean modelos de visión por computadora para la detección de deslizamientos a partir de imágenes satelitales.
Desde el año 2020, se ha registrado un incremento significativo en el número de investigaciones dedicadas a esta área, impulsado por avances tanto en la precisión como en la eficiencia de las técnicas de aprendizaje profundo, así como la creciente disponibilidad de datos satelitales de libre acceso.
Aunque la erosión fluvial y los deslizamientos de tierra son procesos distintos, ambos fenómenos comparten características similares e imágenes satelitales como insumo fundamental. 
Por este motivo, resulta de gran utilidad para nuestro caso de estudio el análisis de los artículos de deslizamientos.
De cada trabajo, se han extraído los aspectos más relevantes, como el objetivo, la metodología empleada, los principales resultados obtenidos, las limitaciones identificadas y las conclusiones derivadas.

\citep{Jichao} se plantean detectar deslizamientos con un modelo de segmentación semántica semisupervisado. Utilizan el entrenamiento adversarial virtual (VAT) para aumentar la robustez del modelo. Crearon un dataset de 1022 imágenes etiquetadas y 2776 no etiquetadas de 512 x 512 píxeles, a partir de imágenes de teledetección por satélite Planet (resolución de 4.77 m) y DEM (resolución de 30 m). Adoptaron un mecanismo de aprendizaje contrastivo a nivel de píxel, estableciendo así un espacio de características de píxeles altamente estructurado.
Los resultados experimentales muestran que el modelo detectó 432 deslizamientos en el área de estudio con una precisión de 0.913, una puntuación F1 de 0.91 y una intersección media sobre unión (mIoU) del 84.2\%. 
A pesar de esto, presenta limitaciones en detectar deslizamientos más pequeños, debido a la resolución espacial insuficiente para detectar estos objetos.  

\citep{Yang} proponen un modelo YOLO ligero a partir de YOLOv4, guiado por atención con una capa de nivelación para la detección precisa de los límites de un deslizamiento. Realizaron un estudio comparativo para seleccionar una estructura de extracción de características óptima a partir de múltiples CNNs. Luego, presentan un nuevo mecanismo de atención y lo incluyen al final de la estructura de YOLO. Por último, combinan YOLO y el conjunto de niveles para extraer el límite preciso del deslizamiento. Utilizaron un conjunto de 770 imágenes sobre deslizamientos y 2100 imágenes que no son de deslizamientos con tamaños de 128 x 128 y 1024 x 1024. Posteriormente, realizan un aumento de imágenes de deslizamientos para equilibrar el dataset. El método propuesto logró una precisión del 95.54\%, una recuperación del 94.29\%, una puntuación F1 del 94.91\% y un mAP del 96.02\%. La capa de nivelación integrada en YOLO mejoró la precisión de la detección de límites de deslizamientos individuales, lo que resultó en un mejor desempeño en la detección tanto del número como del área de deslizamientos.

\citep{Zhang} generan un modelo de detección de deslizamientos (LS-YOLO) a partir de YOLOv5. El módulo de extracción multiescala (MSFE) obtiene características de deslizamientos de varios campos receptivos mediante cinco ramas paralelas. Estas ramas consisten en agrupamiento promedio o convolución espacial separable, lo que aumenta la profundidad de la red y mejora la precisión de detección de deslizamientos del modelo. 
Utilizaron 9,620 imágenes de deslizamientos. Las etiquetas del conjunto de datos de deslizamientos se utilizan para la segmentación de instancias. La precisión de LS-YOLO para la detección de deslizamientos es del 97.60\%, la recuperación es del 94.56\% y la AP es del 97.06\%. LS-YOLO muestra una ventaja considerable tanto en la detección de deslizamientos individuales como de deslizamientos múltiples en imágenes de teledetección en comparación con los modelos existentes. Algunas limitaciones son: una estructura más compleja y una velocidad de detección más lenta.

\citep{Zheng} crean Dynahead-YOLO-Otsu, un método conveniente de segmentación semántica de deslizamientos basado en redes neuronales convolucionales profundas (DCNN).
Este modelo es una versión mejorada de YOLOv3 que consta de tres componentes esenciales: una columna vertebral para extraer características de deslizamientos en la imagen, un cuello para la fusión de características y una cabeza de detección para la localización de deslizamientos. 
El conjunto de datos consta de 1200 imágenes de deslizamientos válidas y visibles, todas ellas con datos de tres canales (RGB) y etiquetas de cuadro delimitador.
El modelo propuesto logró un valor de IoU de 0.65\% y una precisión del 79.47\% para la identificación de deslizamientos de tierra, aproximadamente un 10-30\% mejor que los otros tres modelos de segmentación semántica por píxel. Los autores se encontraron con una limitación en la obtención de imágenes, especialmente en un área extensa.

\citep{Noe} tienen como objetivo utilizar redes neuronales convolucionales (CNNs) para detectar y mapear regiones de ``Brain Terrain'' en Marte, que son formaciones geológicas posiblemente relacionadas con ciclos de congelación y descongelación. Se emplearon imágenes de alta resolución del orbitador Mars Reconnaissance Orbiter (MRO), analizando 58,209 imágenes (28 TB de datos). La metodología combinó una red clasificadora en el dominio de Fourier y una red de segmentación para mejorar la eficiencia computacional, logrando un 0.93 de precisión y reduciendo el tiempo de procesamiento en un 0.95. Se identificaron 201 imágenes con detecciones confirmadas y 1,141 candidatas, lo que sugiere una distribución preferencial en latitudes medias (40°N y 40°S), que podría estar vinculada a condiciones climáticas pasadas en Marte.

La revisión del trabajo existente proporciona una visión general del ámbito en el que los autores han enfocado sus esfuerzos para aplicar la inteligencia artificial. La mayoría de los estudios fueron desarrollados recientemente (2024), con el objetivo de conseguir la detección y delimitación de deslizamientos. El modelo más destacado utilizado es YOLO, al cual se le aplican diversas modificaciones para mejorar su rendimiento. Los conjuntos de datos de imágenes empleados oscilan entre 1200 y 9600 imágenes, donde los modelos propuestos tienen una precisión que varía entre el 80\% y el 97\%. Los principales resultados alcanzados son el mejoramiento del modelo de detección de deslizamientos a partir de las diferentes versiones de YOLO. No obstante, se identifican limitaciones, como la dificultad para detectar deslizamientos de menor escala debido a la resolución de las imágenes satelitales utilizadas, así como los desafíos para obtener dicho tipo de imágenes.

El análisis de imágenes mediante visión por computadora ha experimentado un avance significativo gracias al desarrollo del deep learning. Sin embargo, los estudios existentes se limitan a modelos diseñados exclusivamente para la detección de deslizamientos. En contraste, el presente estudio se enfoca en la detección automática de la erosión fluvial, un fenómeno natural que aún no ha sido abordado desde la perspectiva de la visión por computadora. Además, destacamos que el modelo propuesto también permite estimar el área erosionada. A diferencia de los estudios previos, presentamos una novedosa aplicación web, la cual estará disponible de manera gratuita para una amplia variedad de usuarios.

\section{Metodología}
\label{sec:method}
\begin{figure}[!htbp]
    \centering
\begin{tikzpicture}[node distance=1.75cm]
\node (start) [process] {Recopilación y preprocesamiento de imágenes};
\node (process1) [process, below of=start] {Etiquetado de imágenes (Roboflow)};
\node (process2) [process, below of=process1] {Preparación y descarga del dataset};
\node (process3) [process, below of=process2] {Arquitectura de la solución};
\node (process4) [process, below of=process3] {Entrenamiento};
\node (process5) [process, below of=process4] {Evaluación del modelo};
\node (process6) [process, below of=process5] {Aplicación web};
\draw [arrow] (start) -- (process1);
\draw [arrow] (process1) -- (process2);
\draw [arrow] (process2) -- (process3);
\draw [arrow] (process3) -- (process4);
\draw [arrow] (process4) -- (process5);
\draw [arrow] (process5) -- (process6);

\draw [decorate,decoration={brace,amplitude=5pt,mirror,raise=4ex}]
  (2,-4.25) -- (2,0.75) node[midway,xshift=4.5em]{Datos};
\draw [decorate,decoration={brace,amplitude=5pt,mirror,raise=4ex}]
  (-2,-4.5) -- (-2,-9.5) node[midway,xshift=-5em]{Modelo};
  
\end{tikzpicture}
   
    \caption{Diagrama de flujo de la metodología de trabajo.}
    \label{fig:workflow}
\end{figure}
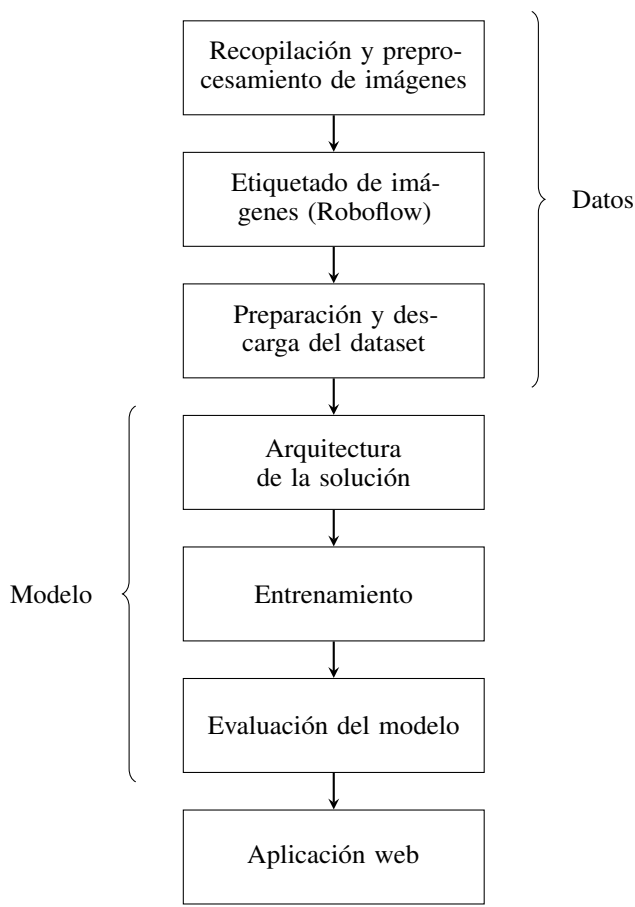

El objetivo de esta investigación es desarrollar un modelo de visión artificial, basado en deep learning, para la detección automática de la erosión fluvial. Este enfoque busca optimizar la identificación de patrones de erosión y zonas de mayor riesgo.
Para tal propósito, se siguió el flujo de trabajo representado en la Figura \ref{fig:workflow}. Básicamente, este flujo se compone de dos grandes etapas. La primera está dedicada a los datos, donde se crea y prepara adecuadamente el dataset de imágenes, que servirá como insumo principal para la siguiente etapa, la cual consiste en entrenar y evaluar el modelo de YOLOv11 para obtener como resultado la identificación de la erosión fluvial.
Una de las contribuciones destacadas de este trabajo es el desarrollo de una aplicación web, donde se despliega el modelo entrenado con el fin de detectar automáticamente las zonas de erosión en las imágenes proporcionadas por el usuario, además de la estimación del área correspondiente.
A continuación, se presentan los detalles de cada una de estas etapas.

\subsection{Recopilación y preprocesamiento de imágenes}
El primer paso fue la recopilación del conjunto de datos, considerando que debe estar compuesto por imágenes que representen claramente el fenómeno de erosión fluvial.
En este estudio, se compiló un conjunto de imágenes LiDAR y fotogramétricas relacionadas con el proceso erosivo del río Coca en la región oriental del Ecuador.
CELEC EP, a través de la Comisión Ejecutora Río Coca (CERC) \citep{Comision}\citep{Comision2}\citep{EGCConsulting} \citep{Gibson}, proporcionó 20 imágenes LiDAR en formato ECW, capturadas entre 2021 y 2024.
Este conjunto de imágenes fue complementado con 40 imágenes en formato JPG, 30 de las cuales fueron recopiladas de la web (noticieros, redes sociales, blogs) y 10 fotografías que fueron tomadas por los autores en el sitio del fenómeno \citep{Primicias}\citep{IIGE}.

Las imágenes LiDAR tienen un peso aproximado de 100 MB cada una y una cobertura de 1.5 km × 1.5 km.
De tal modo, estas imágenes son difíciles de procesar, por lo que se requiere de un preprocesamiento que consiste en dividir las imágenes en fragmentos más pequeños para facilitar su manejo y análisis.
Además, esto permitirá entrenar modelos de manera más eficiente para tareas como la detección y segmentación de objetos.

\begin{figure}[!htb]
    \centering
    \includegraphics[width=0.95\textwidth]{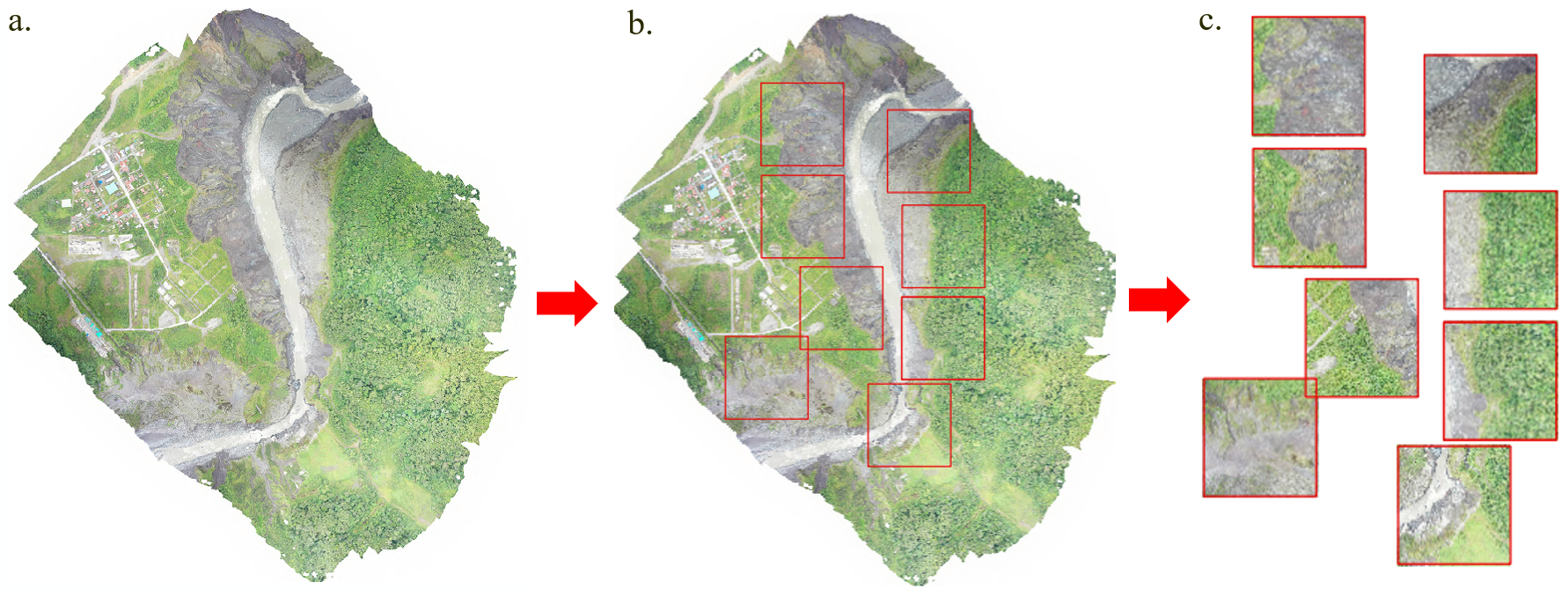}
    \caption{Preprocesamiento de una imagen LiDAR de la erosión del río Coca. a) Imagen original que cubre 1.5 km x 1.5 km, b) zonas identificadas de erosión de 250 m x 250 m, y c) imágenes recortadas.}
    \label{fig:imglidar}
\end{figure}

La Figura \ref{fig:imglidar} ilustra el procedimiento seguido para dividir las imágenes originales en tamaños más pequeños de 250 m × 250 m mediante un sistema de información geográfica (GIS).
En cada imagen, se identificaron las zonas con evidencias más claras de erosión fluvial, las cuales fueron delimitadas, recortadas y convertidas a formato JPG, lo que redujo su peso a un rango entre 100 y 300 KB, optimizando el tiempo de su procesamiento. Como resultado, se generaron 160 imágenes a partir de las 20 imágenes originales.

\begin{figure}[!htb]
    \centering
    \includegraphics[width=0.95\textwidth]{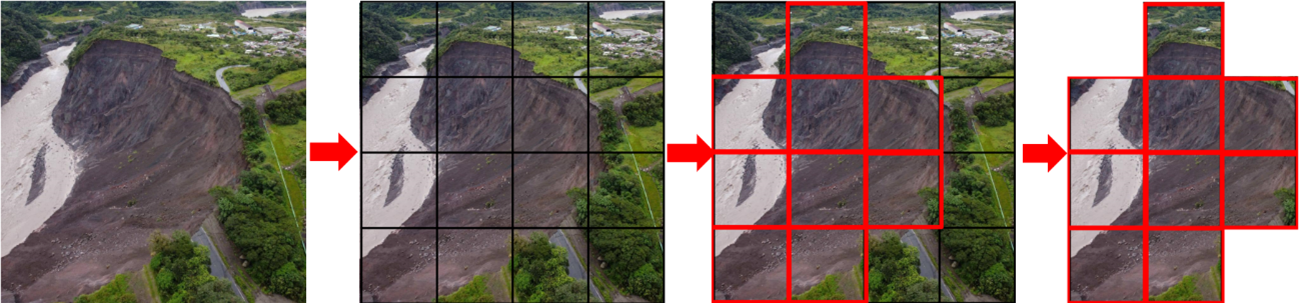}
    \caption{Proceso de recorte de una imagen JPG de la erosión del río Coca.}
    \label{fig:imglidarsplit}
\end{figure}

En cuanto a las imágenes recopiladas de la web y las fotografías de campo, estas fueron procesadas en \textit{Google Colab} mediante código Python, dividiendo la imagen en secciones iguales y seleccionando las áreas más relevantes (Figura \ref{fig:imglidarsplit}). A partir de las 40 imágenes iniciales, se obtuvieron 60 fragmentos. 

Tras la recopilación y recorte de las fotografías, el dataset quedó conformado por un total de 220 imágenes del fenómeno de erosión fluvial.

\subsection{Etiquetado de imágenes (Roboflow)}

Esta labor es esencial en el aprendizaje automático de tipo supervisado, pues el modelo se entrena
conociendo tanto los datos de entrada (imágenes) como las respuestas, es decir, las etiquetas que le dicen al modelo qué hay en la imagen.
De este modo, el modelo puede aprender el mapeo de las entradas a las salidas correspondientes.

El etiquetado o anotación es un proceso que exige tiempo, precisión y conocimientos del dominio del problema.
Dependiendo del objetivo, las etiquetas tienen formatos específicos.
En la detección de objetos, las etiquetas suelen ser cajas o cuadros (bounding boxes) alrededor del objeto, junto con la anotación de la clase respectiva.
En la segmentación, la forma del objeto debe ser delimitada con exactitud, por ejemplo, a través de polígonos o máscaras.

En ambos casos, pueden presentarse diferentes tamaños, formas, orientaciones, superposiciones y oclusiones de los objetos dentro de la imagen.
Además, si las imágenes tienen distintas propiedades de tamaño, resolución, color, iluminación y formato, y la cantidad es significativa, se requiere de mucho esfuerzo y criterio del experto.

Por tanto, el  etiquetado no es una tarea fácil y demanda una herramienta especializada que reduzca la complejidad de llevarla a cabo.
\textit{Roboflow} \citep{roboflow} es una plataforma para gestionar y etiquetar imágenes muy conocida en visión por computadora.
Permite la segmentación manual o asistida por IA para entrenar modelos \citep{Alexandrova}.

Los pasos a seguir para el etiquetado en Roboflow son: 

\begin{itemize}
    \item Creación de un nuevo proyecto.
    
    \item Carga del conjunto de imágenes subiéndolas al proyecto. Roboflow acepta múltiples formatos, puede organizar las imágenes por lotes o carpetas, y permite mantener un control de versiones del dataset.

    \item Anotación manual o asistida: Las imágenes se etiquetan utilizando herramientas integradas para dibujar cajas delimitadoras (bounding boxes), polígonos o máscaras, según el tipo de tarea. También es posible usar anotaciones automáticas asistidas por modelos preentrenados o usar IA para acelerar el proceso.

     \item Revisión y ajuste de etiquetas: Una vez generadas las anotaciones, se revisan y ajustan para garantizar la calidad y precisión de los datos etiquetados. Esto es crucial en tareas complejas como la segmentación semántica o la detección con múltiples objetos superpuestos.

     \item Preprocesamiento y aumentos de datos (data augmentation): Antes de exportar el dataset, Roboflow permite aplicar técnicas de aumento de datos, como rotación, cambio de escala, ajuste de brillo, entre otros. Esto ayuda a mejorar la robustez del modelo durante el entrenamiento.
     
     \item Exportación del dataset en el formato compatible con el framework de entrenamiento que se utilizará (YOLO, TensorFlow, COCO JSON, etc.).
\end{itemize}


En este estudio, se aprovechó un modelo preentrenado con un dataset previamente disponible en Roboflow conformado por 849 imágenes de deslizamientos segmentadas por instancias. A este dataset base, se añadieron las nuevas imágenes LiDAR cortadas y propias de este proyecto. Este dataset extendido es útil para el etiquetado automático.

Las etiquetas van a clasificar los objetos según el contorno que tenga cada uno.
Si bien la clase de interés es la erosión fluvial, se integraron dos clases adicionales: `río' (cuerpo de agua principal) y `aluvial' (depósito de materiales transportados por el agua), es decir, elementos geográficos vecinos, pero diferentes. Esta estrategia pretende un aprendizaje más distintivo en relación al entorno y delimitar con precisión la clase principal.
En un inicio, se consideraron estas tres clases, pero al etiquetar en Roboflow ciertas imágenes presentan suelo y vegetación, lo que causaba confusión al modelo. Por tal razón, al final se integraron estas dos nuevas clases para una mejor comprensión y señalización del área de interés.
La delimitación de estas clases se realizó mediante dos métodos: manual y automático.

\begin{figure}[!htb]
\centering
\includegraphics[width=0.32\textwidth]{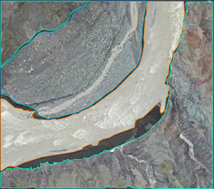}
\includegraphics[width=0.5\textwidth]{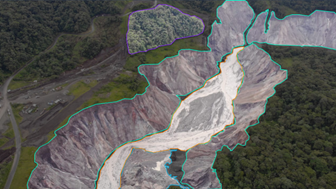}
    \caption{Imágenes etiquetadas por categorías. Erosión fluvial (polígono verde), aluvial (celeste) y río (naranja).}
    \label{fig:categorysample}
\end{figure}

El etiquetado manual utiliza la herramienta de anotación de polígonos, siguiendo puntos individuales en la imagen para delimitar las áreas de interés. En contraste, la herramienta IA de detección de Roboflow genera polígonos de manera asistida. El modelo de visión por computadora \textit{Segment Anything Model} (SAM) es capaz de etiquetar con gran precisión con unos pocos clics, eliminando el tedioso proceso manual de seleccionar cada punto.
Este enfoque aceleró considerablemente el proceso de anotación. Finalmente, cada polígono generado fue clasificado en su categoría correspondiente (Figura \ref{fig:categorysample}).


La organización del dataset resultante se estructura en dos carpetas:
``images'' y ``labels'', cada una con sus respectivas subcarpetas de train, val y test para las tres categorías citadas anteriormente.
Los archivos de etiquetas siguen el formato YOLO, donde cada línea representa un objeto detectado con la siguiente estructura: \textit{clase}, \textit{xcentro}, \textit{ycentro}, \textit{ancho} y \textit{alto}. Todos los valores están normalizados entre 0 y 1.

\subsection{Preparación y descarga del dataset}
Una vez que las imágenes han sido etiquetadas, en esta fase se realizan las actividades necesarias para disponer de un dataset preparado de manera conveniente para el proceso de entrenamiento del modelo de aprendizaje automático. Estas actividades involucran la depuración, aumento artificial, división y descarga del dataset en el formato adecuado.

El dataset original de 849 imágenes, perteneciente al modelo de detección de deslizamientos, fue depurado y reducido a 280 imágenes. Esta depuración consistió en eliminar imágenes repetidas, de baja calidad o sin presencia clara de deslizamientos. A este conjunto se añadieron las 220 imágenes de erosión fluvial ya etiquetadas, obteniendo un dataset total de 500 imágenes.

Para ampliar el dataset, se aplicaron técnicas de aumento artificial de datos. Las transformaciones incluyen imágenes en blanco y negro, zoom y rotación de 90°. Estas transformaciones incrementan la diversidad del conjunto y mejoran la robustez del modelo, lo que incrementó el dataset a 1204 imágenes.

La división del dataset se realizó de manera aleatoria, asignando porcentajes de 88\% de las imágenes al conjunto de entrenamiento, 6\% al de validación y 6\% al de prueba. Se verificó el balance de clases dentro del conjunto de datos. Aunque algunas clases, como vegetación, presentan menor representación, el aumento de datos permitió mitigar en parte este desequilibrio.

La Tabla \ref{tab:dataset} resume la evolución y distribución del dataset generado. Finalmente, el dataset fue descargado en formato ZIP con el modelo de segmentación de instancias YOLOv11 y punto de control COCOs-seg.

\begin{table}[!htb]
    \centering
    \caption{Distribución y cantidad de imágenes para el dataset}
    \label{tab:dataset}
    \renewcommand{\arraystretch}{1.5} 
    \setlength{\arrayrulewidth}{0.3mm} 
    \begin{tabular}{lc}
        \toprule
        \textbf{Datos} & \textbf{Imágenes} \\ 
        \midrule
        Preentrenado & 849 \\ 
        Depurado (d) & 280 \\ 
        Img. de erosión (ie) & 220 \\ 
        Dataset final (d + ie) & 500 \\ 
        Dataset (con aumento) & 1204 \\ 
        Entrenamiento & 1056 \\ 
        Validación & 74 \\ 
        Test & 74 \\ 
        \bottomrule
     \end{tabular}
\end{table}

\subsection{Arquitectura de la solución} 

\begin{figure}[!htb]
\centering
\includegraphics[width=0.99\textwidth]{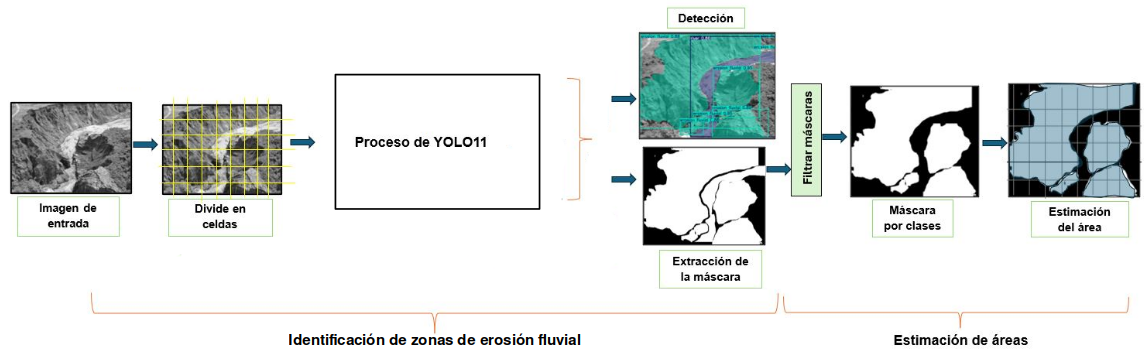}
    \caption{Arquitectura de la solución: detección con YOLOv11 y cálculo de área de erosión fluvial.}
    \label{fig:architecturesolution}
\end{figure}

Una vez preparado el dataset de imágenes necesario para el aprendizaje automático, esta sección explica la solución diseñada para la problemática. El esquema general de su arquitectura puede observarse en la Figura \ref{fig:architecturesolution}.
El componente central es YOLOv11, que se encarga de extraer las características relevantes de la imagen de entrada, las cuales son procesadas para clasificar, localizar y segmentar los objetos dentro de la imagen.
Las funciones de clasificación y localización permiten la identificación de zonas de erosión, mientras que la segmentación es útil para obtener la estimación del área afectada. Tanto el modelo de YOLO como el mecanismo de cálculo del área se detallan a continuación. 

\subsubsection{Detección con YOLOv11} 

El algoritmo \textit{YOLO} (You Only Look Once), basado en una red neuronal convolucional, fue introducido por \citep{Redmon} para detectar objetos en imágenes. El nombre refleja el enfoque utilizado para analizar una sola vez toda la imagen, en lugar de hacerlo en varios pasos como los métodos convencionales \citep{yolov5}. 
De ahí se desprenden sus principales ventajas, como la velocidad y la capacidad de procesar imágenes en tiempo real. Además, se destaca la gran precisión, así como la flexibilidad y adaptabilidad que facilitan su implementación en una amplia variedad de entornos, ya sea en sistemas locales o en la nube \citep{Ultralytics}.


Su rápida evolución ha ido incorporando mejoras significativas con cada versión, expandiendo el rango de tareas que se pueden realizar con imágenes. Esto ha permitido que se mantenga a la vanguardia en el área de la visión por computadora.
Actualmente, la empresa Ultralytics ofrece el sistema de procesamiento de imágenes denominado \textit{YOLOv11} que contiene la detección de objetos, segmentación de instancias, segmentación semántica, estimación de pose, entre otras tareas.
En este proyecto, interesan especialmente dos de ellas: la detección y la segmentación.
Por una parte, la detección de objetos identifica y localiza objetos en la imagen mediante cajas delimitadoras. Por otro lado, la segmentación va un paso más allá y, como su nombre sugiere, se refiere a delinear la forma exacta de cada objeto y segmentarlos del resto de la imagen.

\begin{figure}[!htb]
\centering
\includegraphics[width=0.7\textwidth]{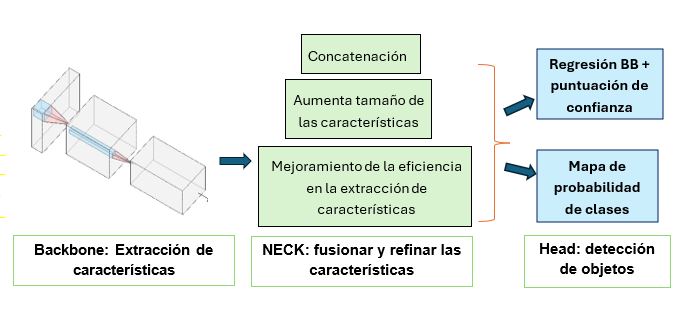}
   
    \caption{Componentes de la arquitectura de YOLOv11.}
    \label{fig:yolov11}
\end{figure}

Por tanto, la detección de objetos es útil para identificar y localizar las zonas de erosión fluvial dentro de una imagen, mientras que la segmentación proporciona una delimitación más precisa de su contorno y forma con el fin de calcular el área ocupada. Para este propósito, se explican a continuación los tres componentes fundamentales de la arquitectura de YOLOv11, la cual se ilustra en la Figura \ref{fig:yolov11} \citep{Rao}\citep{yolov11}\citep{Boesch}\citep{Seong}. 

\begin{itemize}
    \item \textbf{Backbone}: es el responsable de extraer características de la imagen de entrada mediante filtros. La imagen de tamaño 640x640x3 es previamente dividida en una cuadrícula de celdas para realizar la detección en una sola pasada.
    Una CNN procesa estas celdas reduciendo gradualmente sus dimensiones espaciales mientras se incrementa el número de canales.
    Como resultado, se generan mapas de características en diferentes escalas.
    \item \textbf{Neck}: tiene como objetivo mejorar la representación de características antes de la detección final.
    Primero se incrementa la resolución espacial de los mapas de características multiescala. Luego se concatenan para combinar diferentes escalas y captar objetos de distintos tamaños. Esta información es transmitida a la fase de predicción.    
    \item \textbf{Head}: genera las predicciones de la red con base en los mapas de características. Estas predicciones son de dos tipos: regresión de las coordenadas de las cajas delimitadoras de los objetos en la imagen y clasificación del objeto en las categorías consideradas.
    También se obtiene un puntaje de confianza que indica si hay un objeto en la caja delimitadora. Se añade una rama adicional en la cabeza de detección que genera máscaras de segmentación dentro de cada caja delimitadora que contiene el contorno del objeto. Este último resultado ha sido aprovechado para cuantificar el área de los cambios erosivos.

    
\end{itemize}



\subsubsection{Cálculo del área de erosión fluvial} 

El método empleado para el cálculo del área afectada por la erosión fluvial está basado en las máscaras generadas por YOLO durante la fase de segmentación.
El modelo genera mapas de probabilidad de clases, que contienen la segmentación de diferentes categorías dentro de la imagen de entrada. El procedimiento incluye los siguientes pasos (Figura \ref{fig:estimation}):

\begin{figure}[!htb]
\centering
\includegraphics[width=0.95\textwidth]{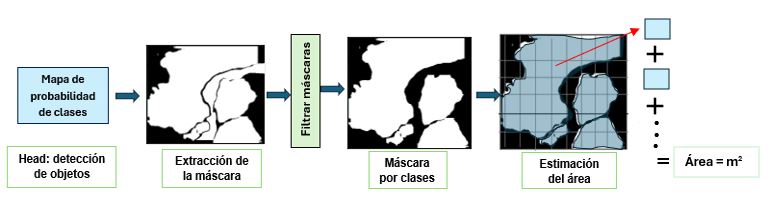}    
    \caption{Filtrado de máscaras y estimación del área de erosión fluvial.}
    \label{fig:estimation}
\end{figure}

\begin{enumerate}
    \item Extraer únicamente las máscaras generadas correspondientes a la categoría de erosión fluvial. Para tal fin, se filtra y descarta aquellas que pertenecen a otras categorías mediante un código en lenguaje Python.
    \item Con las máscaras filtradas, el cálculo está basado en el concepto de que una imagen está conformada por un cierto número de píxeles, y cada uno de estos tiene un tamaño definido \citep{Esri}. Aquí se permite al usuario elegir el resultado entre unidades de píxeles o en metros cuadrados.
    \item Al final se obtiene el área de cada píxel y la suma de estas áreas resulta en un área total de la erosión detectada en la imagen. Se utilizan librerías como NumPy y OpenCV; por ejemplo, la función \textit{cv2.countNonZero()} cuenta los píxeles blancos en la máscara binaria de erosión, y luego se multiplica por el tamaño de píxel definido por el usuario. Si un píxel representa 1 m², entonces 500 píxeles blancos equivalen a 500 m².
    
\end{enumerate}

De esta manera, se ha presentado un enfoque que permite automatizar y optimizar el análisis de erosión fluvial, aprovechando las capacidades de detección y segmentación de YOLOv11, junto con un método cuantitativo de estimación del área.

\section{Experimentación y resultados}
\label{sec:experimental}
Esta sección explica la parte experimental llevada a cabo para obtener un modelo preciso y eficiente que detecte la erosión fluvial y a la vez calcule su área. 
Se describe la plataforma computacional utilizada y los procesos de entrenamiento, evaluación y estimación del área, así como la presentación de sus respectivos resultados.

\subsection{Plataforma computacional} 

Un entorno apropiado de hardware y software es clave para el éxito de un proyecto de inteligencia artificial.
Aunque se realizaron pruebas en un entorno local con procesador Intel I7, 16 GB de RAM y GPU NVIDIA RTX 4050 6 GB, se presentaron problemas como el tiempo de ejecución elevado y errores por falta de recursos.

Se optó por el uso de \textit{Google Colaboratory}, una plataforma en la nube de acceso libre y que permite la ejecución de código desde un navegador web, aprovechando poderosos recursos computacionales remotos.
Facilita la integración con librerías como \textit{PyTorch} y \textit{TensorFlow}, y es compatible con \textit{Google Drive} para el almacenamiento de datos.

Las características técnicas del hardware incluyen un procesador Intel Xeon, 12.7 GB de RAM y una GPU NVIDIA Tesla
T4 con 15 GB de RAM.
En cuanto al software, el sistema operativo Linux, el lenguaje de programación Python 3.10, un notebook para la implementación de \textit{Ultralytics YOLOv11} (detección y segmentación de objetos con modelos preentrenados) y el soporte de librerías como \textit{OpenCV} (procesamiento de imágenes), \textit{NumPy} (operaciones matriciales y análisis numérico) y \textit{YAML} (estructurar archivos de configuración necesarios en el entrenamiento del modelo).



\subsection{Entrenamiento}

El entrenamiento del modelo es la etapa crucial del aprendizaje automático. Es un proceso iterativo donde el algoritmo ajusta sus parámetros para minimizar el error en las predicciones, aprendiendo patrones utilizando los datos suministrados como entrada. En este caso, el objetivo es que el modelo aprenda a predecir de manera precisa la clase y localización (detección), así como la forma de los objetos dentro de las imágenes (segmentación) a partir del conjunto de datos etiquetados.

Para mejorar la precisión en la detección y segmentación de la erosión fluvial, se aplicó \textit{Fine-tuning}. Esta técnica reutiliza el conocimiento adquirido por un modelo ya entrenado para adaptarlo a una tarea específica, evitando la necesidad de entrenar el modelo desde cero.
Con el fin de optimizar el rendimiento del modelo y el tiempo de cómputo, se empleó la versión \textit{nano} de YOLOv11n-seg, que tiene el menor número de parámetros (2.6 M aprox.).
Este modelo ha sido previamente entrenado con un conjunto de datos genérico y extenso para realizar simultáneamente las tareas de detección de objetos mediante \textit{bounding boxes} y segmentación de instancias mediante \textit{máscaras}.
En esta etapa, el modelo es entrenado con el nuevo conjunto de imágenes y etiquetas más específico, ajustando los últimos parámetros del modelo, el cual aprende características particulares relacionadas con las nuevas clases definidas (erosión fluvial, río, aluvial, suelo y vegetación).

El recurso esencial para el entrenamiento son los datos. De acuerdo con la Tabla \ref{tab:dataset}, el conjunto de datos está conformado por 1204 imágenes con sus respectivas etiquetas. De este total, 1056 se asignan para el entrenamiento del modelo, 74 para la validación y 74 para la prueba. En la carpeta principal se crean dos subcarpetas: una para las imágenes y otra para las etiquetas. Dentro de cada subcarpeta, se crean las carpetas de train, val y test. Las imágenes y etiquetas de la carpeta train se usan para entrenar y afinar el modelo. En cada iteración del entrenamiento o afinamiento, el desempeño del modelo es evaluado con las imágenes de la carpeta de val, y las imágenes de la carpeta test sirven para la prueba final del modelo con imágenes no vistas.

Antes del entrenamiento, es necesario un índice para la codificación de las distintas clases a detectar mediante la creación de un archivo \textit{.yaml}. Se deben especificar las rutas del dataset y de las subcarpetas: train, val y test, así como también las 5 clases consideradas: 0: 'suelo', 1: 'vegetación', 2: 'aluvial', 3: 'erosión fluvial' y 4: 'río'.
Una vez organizado el conjunto de datos, se deben configurar los \textit{hiperparámetros}, que son valores que buscan garantizar un entrenamiento eficiente y estable (Tabla \ref{tab:yolo_hyperparams}).

\begin{table}[htb!]
\centering
\caption{Hiperparámetros del entrenamiento.}
\label{tab:yolo_hyperparams}
\scalebox{0.9}{
\begin{tabular}{lll}
\toprule
\textbf{Hiperparámetro} & \textbf{Descripción} & \textbf{Valor} \\
\midrule
\textbf{image\_size} & Tamaño de imagen & 640x640 px. \\
\textbf{channels} & Número de canales de color & 3 (RGB) \\
\textbf{epochs} & Iteraciones del conjunto de datos & 50 \\
\textbf{optimizer} & Algoritmo de optimización & adamw \\
\textbf{lr0} & Tasa de aprendizaje inicial & 0.01 \\
\textbf{momentum} & Momentum para el optimizador & 0.937 \\
\textbf{weight\_decay} & Decaimiento de peso para regularización L2 & 0.0005 \\
\textbf{warmup\_epochs} & Épocas de calentamiento para el learning rate & 3 \\
\textbf{warmup\_momentum} & Momentum durante el calentamiento & 0.8 \\
\textbf{warmup\_bias\_lr} & Tasa de aprendizaje para el bias durante el calentamiento & 0.1 \\
\textbf{box} & Peso de la pérdida del bounding box & 0.05 \\
\textbf{cls} & Peso de la pérdida de clasificación & 0.5 \\
\textbf{cls\_pw} & Peso de la pérdida de clasificación positiva & 1.0 \\
\textbf{obj} & Peso de la pérdida de confianza del objeto & 1.0 \\
\textbf{obj\_pw} & Peso de la pérdida de confianza del objeto positiva & 1.0 \\
\textbf{iou\_t} & Umbral de IoU para la supresión no máxima (NMS) & 0.2 \\
\textbf{anchor\_t} & Umbral de anclaje para la asignación de objetivos & 4.0 \\
\textbf{hsv\_h} & Aumento de datos: variación del tono (hue) & 0.015 \\
\textbf{hsv\_s} & Aumento de datos: variación de la saturación (saturation) & 0.7 \\
\textbf{hsv\_v} & Aumento de datos: variación del valor (value) & 0.4 \\
\textbf{translate} & Aumento de datos: traslación de la imagen & 0.1 \\
\textbf{scale} & Aumento de datos: escala de la imagen & 0.5 \\
\textbf{flip\_lr} & Aumento de datos: volteo horizontal & 0.5 \\
\textbf{mosaic} & Aumento de datos: uso de mosaico & 1.0 \\
\bottomrule
\end{tabular}
}
\end{table}

El proceso de entrenamiento se lleva a cabo de la siguiente manera:

\begin{enumerate}
    \item Se descarga YOLOv11 del repositorio de Ultralytics con los parámetros preentrenados.
    \item Se leen las imágenes por lotes desde la carpeta de entrenamiento y pasan a través de la red, la cual genera predicciones de la clase, las coordenadas de las cajas delimitadoras, la confianza de que existe un objeto y la máscara de segmentación.
    \item Se calculan las pérdidas o errores que comete la red. Entre estos valores, destacan la pérdida de clasificación, localización y segmentación. Estos cálculos se realizan tanto para las imágenes de entrenamiento como para las de validación.
    \item Se reducen los valores de pérdida de entrenamiento a través de \textit{backpropagation}, que es una técnica de optimización basada en el algoritmo del descenso de gradiente con respecto a los parámetros de la red.
    \item Se actualizan los parámetros de la red y los pasos 2 al 5 se repiten 50 veces con el fin de minimizar los errores del modelo.
\end{enumerate}

Tras completar el entrenamiento, el modelo YOLO aprende a identificar, localizar y segmentar las zonas de erosión fluvial dentro de las imágenes.
El modelo final alcanzó una precisión de 0.7, valor obtenido directamente del resumen de validación generado por YOLOv11n-seg. Este resultado se considera satisfactorio para la problemática tratada, especialmente debido a la adecuada cobertura del área de interés.

\begin{figure}[!htb]
\centering
\includegraphics[width=0.9\textwidth]{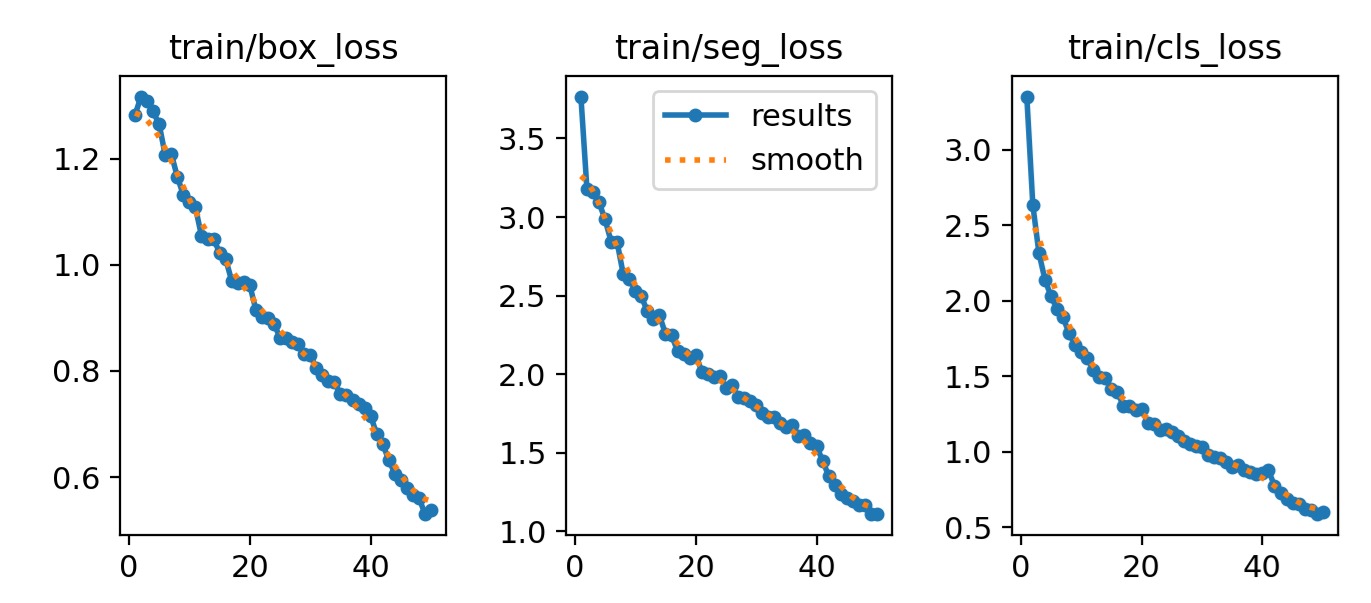}    
    \caption{Curvas de apendizaje del entrenamiento de YOLOv11-seg.}
    \label{fig:learningcurves}
\end{figure}

El proceso de entrenamiento genera diferentes métricas de rendimiento, las cuales son impresas en la consola y guardadas en un archivo de historial (\textit{.log}).
Estas métricas incluyen la pérdida total, pérdidas individuales, precisión, recall y mAP.
Es posible observar estas métricas a través de tablas y gráficas, con el fin de analizar su evolución y el rendimiento del modelo.
La Figura \ref{fig:learningcurves} presenta las \textit{curvas de aprendizaje}, que indican el valor de la pérdida (eje vertical) a lo largo de las épocas o iteraciones (eje horizontal) para el conjunto de imágenes de entrenamiento.
La pérdida de clasificación (\textit{cls\_loss}) calcula la diferencia entre las clases predichas y las clases de referencia; se suele usar la función de entropía cruzada (\textit{Cross-Entropy}).
La pérdida de localización (\textit{box\_loss}) calcula la diferencia entre las coordenadas predichas de la caja delimitadora y las coordenadas de referencia; se suele usar la pérdida de \textit{error cuadrático medio} (MSE) o la pérdida de \textit{intersección sobre unión} (IoU).
La pérdida de segmentación (\textit{seg\_loss}) calcula la diferencia entre las máscaras predichas y las de referencia; se puede usar la pérdida de \textit{Dice} o la pérdida de entropía cruzada para píxeles.

Se puede decir que el comportamiento de las curvas que representan las pérdidas de clasificación, localización y segmentación es aceptable; en todos los casos se aproximan a cero a medida que avanzan las épocas.
Esto indica una mejora progresiva en el desempeño del modelo; es decir, la pérdida va disminuyendo con el tiempo, así que el modelo está aprendiendo correctamente.
También se generaron curvas de validación, las cuales presentan un comportamiento similar al de entrenamiento, lo que sugiere que no hay sobreajuste (overfitting) ni subajuste (underfitting).
El modelo generaliza bien sobre datos no vistos, lo cual es confirmado posteriormente en la evaluación con el conjunto de test.

La Tabla \ref{tab:resultados_modelo} muestra otras métricas que son clave para conocer el desempeño del modelo para cada clase detectada y segmentada (suelo, vegetación, aluvial, erosión fluvial, río y una adicional generada por YOLO para obtener un rendimiento global del modelo llamada ‘all’). 
Se incluyen el número de imágenes en el dataset que contienen al menos una instancia de esa clase, el número total de instancias o áreas anotadas para esa clase, la precisión para detección de cajas delimitadoras, el valor del recall, la \textit{Mean Average Precision} (mAP) a un umbral de IoU de 0.50 y el mAP promediado sobre múltiples umbrales de IoU, desde 0.50 hasta 0.95. Un mayor IoU indica que la predicción del modelo se acerca más a la segmentación real de la erosión fluvial.

\begin{table}[!htb]
    \centering
    \caption{Métricas del modelo sobre el conjunto de datos de validación.}
    \begin{tabular}{lcccccc}
        \toprule
        \textbf{Clase} & \textbf{Imágenes} & \textbf{Instancias} & \textbf{Box(P)} & \textbf{R} & \textbf{mAP50} & \textbf{mAP50-95} \\
        \midrule
        all             & 74  & 178 & 0.618 & 0.524 & 0.550 & 0.373 \\
        suelo          & 35  & 41  & 0.743 & 0.704 & 0.810 & 0.472 \\
        vegetación     & 10  & 12  & 0.285 & 0.332 & 0.247 & 0.190 \\
        aluvial        & 15  & 28  & 0.478 & 0.429 & 0.413 & 0.258 \\
        río & 38  & 64  & 0.697 & 0.395 & 0.502 & 0.352 \\
        \rowcolor{green!15} erosión fluvial            & 26  & 33  & 0.890 & 0.758 & 0.777 & 0.594 \\
        \bottomrule
    \end{tabular}
    \label{tab:resultados_modelo}
\end{table}

El modelo resulta exitoso en la tarea principal de identificar, detectar y estimar áreas de erosión fluvial.
Esta es la clase con el mejor rendimiento, con una precisión muy alta (0.890) que indica que las predicciones de erosión fluvial son casi siempre correctas; un recall alto (0.758) que significa que detecta la mayoría de las instancias de erosión fluvial; un mAP50 muy alto (0.777) y mAP50-95 también muy fuerte (0.594) que demuestran que el modelo no solo detecta la erosión fluvial de manera confiable, sino que también la localiza con alta precisión en las imágenes.

\begin{figure}[!htb]
\centering
\includegraphics[width=0.55\textwidth]{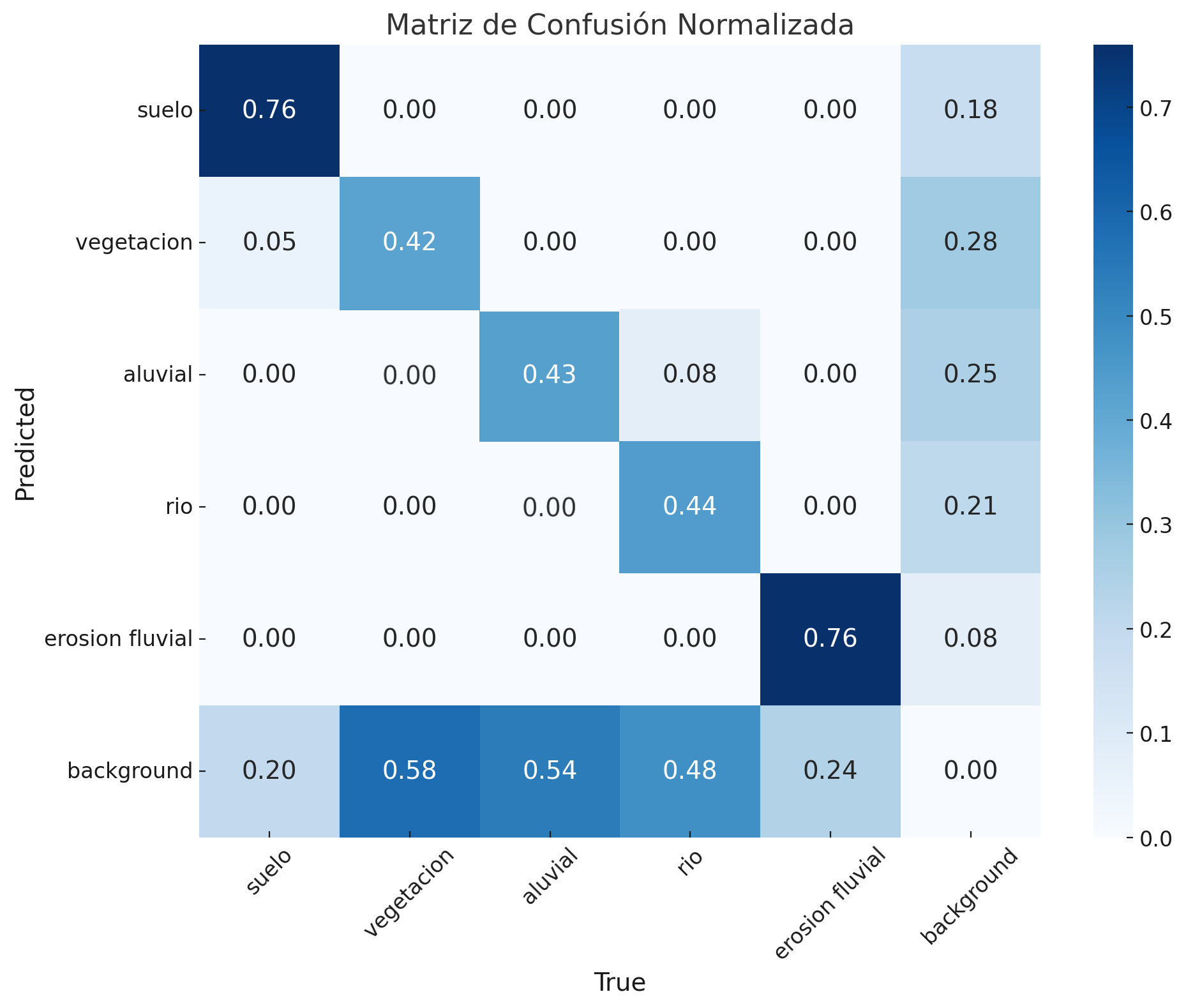}  
    \caption{Matriz de confusión del entrenamiento.}
    \label{fig:confusionmatrix}
\end{figure}

Para complementar el análisis, una herramienta útil es la \textit{matriz de confusión}, la cual permite establecer los aciertos y fallos del modelo para cada clase en particular.
Es una tabla cuyas filas son las clases predichas y las columnas son las etiquetas de verdad. Por ende, la diagonal principal de la matriz contiene los aciertos, mientras que los valores fuera de esta diagonal corresponden a los fallos.
La Figura \ref{fig:confusionmatrix} muestra dichos valores de manera normalizada.

El nivel de aciertos alcanzado por el modelo para la erosión fluvial es bastante aceptable. Es importante destacar que las demás clases fueron añadidas exclusivamente para mejorar la diferenciación con los elementos contextuales.
Por lo tanto, los valores obtenidos para las clases adicionales no tienen una relevancia significativa en el desempeño general del modelo.
Los resultados obtenidos del entrenamiento son muy prometedores para el objetivo central del estudio.

\subsection{Evaluación}
Una vez finalizado el entrenamiento, se procede a evaluar el modelo utilizando el conjunto de datos de prueba (test), el cual está compuesto por 74 imágenes no vistas previamente. Esta etapa permite medir la capacidad del modelo para generalizar sobre nuevos datos y simular su desempeño en escenarios reales.

El modelo utilizado para la evaluación fue el que presentó el mejor rendimiento durante el entrenamiento. Se aplicó la técnica de data augmentation previamente, lo que permitió que el modelo fuera más robusto frente a variaciones en iluminación, escala, resolución o ángulo de las imágenes. Gracias a estas transformaciones, el modelo pudo adaptarse a distintas condiciones geográficas.

La Tabla \ref{tab:resultados_modelo_metricas} muestra los resultados de la evaluación, los cuales son similares a los presentados en la Tabla \ref{tab:resultados_modelo}.
Los valores principales de la clase de mayor interés son aceptables, reflejan un desempeño sólido y una buena capacidad de generalización.

\begin{table}[!htb]
    \centering
    \caption{Métricas del modelo sobre el conjunto de datos de test.}
    \begin{tabular}{lcccccc}
        \toprule
        \textbf{Clase} & \textbf{Imágenes} & \textbf{Instancias} & \textbf{Box(P)} & \textbf{R} & \textbf{mAP50} & \textbf{mAP50-95} \\
        \midrule
        all             & 74  & 209 & 0.479 & 0.548 & 0.48 & 0.328 \\
        suelo          & 36  & 41  & 0.656 & 0.683 & 0.708 & 0.56 \\
        vegetación     & 13  & 19  & 0.308 & 0.351 & 0.193 & 0.082 \\
        aluvial        & 18  & 33  & 0.5 & 0.606 & 0.524 & 0.318 \\
        río & 33  & 76  & 0.481 & 0.525 & 0.461 & 0.282 \\
        \rowcolor{green!15} erosión fluvial & 26  & 40  & 0.689 & 0.675 & 0.714 & 0.694 \\
        \bottomrule
    \end{tabular}
    \label{tab:resultados_modelo_metricas}
\end{table}

\begin{figure}[!htb]
\centering
\includegraphics[width=0.95\textwidth]{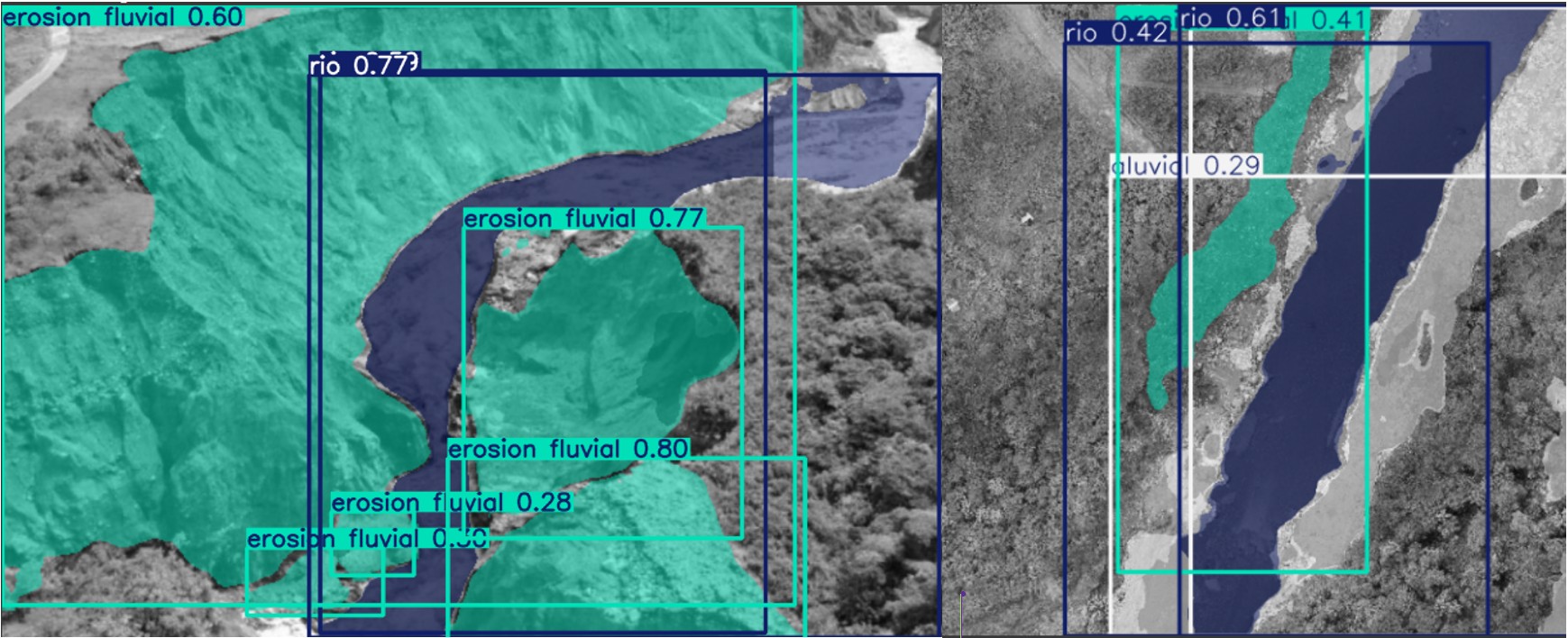}
    \caption{Ejemplos de predicción del modelo con imágenes de test. Se verifica la detección y segmentación de las principales clases: erosión fluvial, río y aluvial.}
    \label{fig:test}
\end{figure}

Aunque el mAP50 presenta una ligera disminución de 0.06 respecto a validación, sigue siendo muy similar y aceptable para aplicaciones de segmentación en entornos naturales, donde la variabilidad de las imágenes es alta.
El mAP50-95 es un indicador más estricto y es recomendable mantenerlo entre 0.4-0.65 según la literatura, lo que indicaría que el modelo logra un equilibrio entre precisión y capacidad de detección en distintos niveles de superposición de objetos.
La leve caída en datos de test indica que el modelo enfrenta nuevos patrones o variaciones en los datos, pero mantiene una precisión aceptable.
La precisión se mantuvo cercana a 0.7, asegurando que la mayoría de las predicciones sean correctas (Figura \ref{fig:test}), mientras que el recall de la clase de erosión mostró un ligero aumento en test, volviéndolo más aceptable para detección de fenómenos geoespaciales complejos.
En cuanto a las pérdidas de clasificación y localización, estas se mantuvieron estables con valores inferiores a 0.05, lo que confirma que el modelo identifica correctamente la ubicación y clasificación de las zonas de erosión, río y aluvial.
Con respecto a la pérdida de segmentación, aunque presentó un leve aumento en test, manteniéndose en el rango de 0.05-0.15, sigue siendo aceptable para tareas de segmentación en imágenes con alta variabilidad geográfica. Un incremento moderado en esta métrica es esperado cuando se aplican modelos a datos nuevos, especialmente en escenarios con iluminación, escala o calidad variable.

En conjunto, tal como se muestra en la Figura \ref{fig:test}, estos resultados confirman que el modelo es funcional para la detección y segmentación de erosión fluvial. Sin embargo, se cree que estos valores pueden mejorar en el futuro, para lo cual se sugiere aumentar la diversidad del conjunto de entrenamiento y optimizar la segmentación sin comprometer la estabilidad de las predicciones.


\subsection{Cálculo del área} 
\begin{figure}[!htb]
\centering
\includegraphics[width=0.75\textwidth]{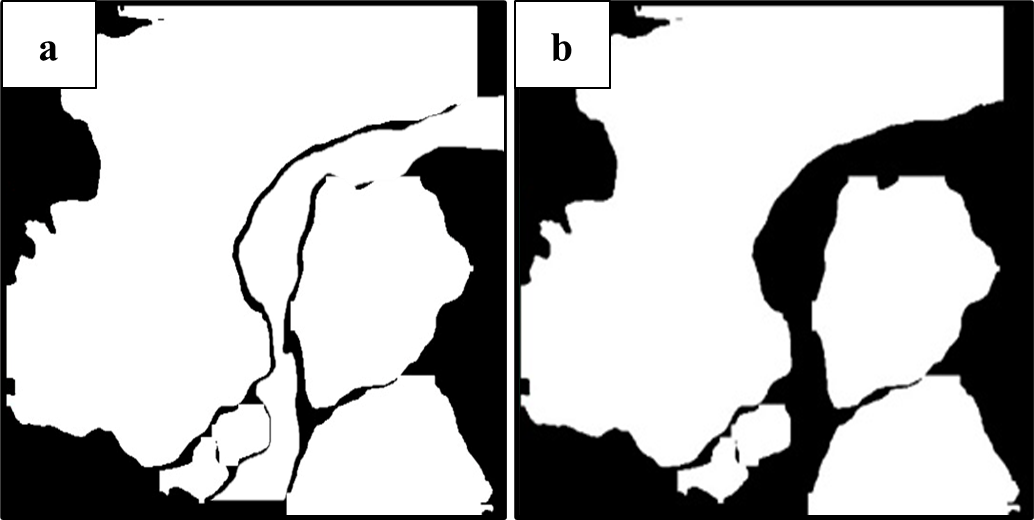}
    \caption{Ejemplo de generación de máscara. a) Máscara con todas las clases de objetos detectados; y b) Máscara con los objetos filtrados de erosión fluvial.}
    \label{fig:masks}
\end{figure}

A partir de la identificación correcta de la erosión fluvial comprobada en la evaluación del modelo, se procedió a la siguiente fase, que consiste en la generación de una máscara de los objetos detectados.
Una vez detectadas las diferentes máscaras, lo blanco corresponde al área erosionada y lo negro a la no erosionada (Figura \ref{fig:masks}). Posteriormente, se aplica un filtro para aislar exclusivamente la máscara correspondiente a la erosión fluvial. 
El cálculo del área de la máscara se realiza mediante la especificación de una unidad de medida por parte del usuario, que puede ser el tamaño del píxel o en metros cuadrados. El modelo está diseñado para procesar ambas entradas de manera intercambiable, asegurando la equivalencia de los resultados obtenidos. Finalmente, la estimación del área se presenta en metros cuadrados (m²) o en píxeles (px).

\section{Discusión} 
\label{sec:discussion}

Los estudios recientes están enfocados en realizar modelos de detección automática para deslizamientos, por ejemplo \citep{Zhang}, \citep{Yang}, \citep{Jichao}, entre otros.  Por esta razón, el presente estudio se enfocó en la detección de la erosión fluvial a través del modelo YOLOv11. Esta versión mejora la extracción de características con una arquitectura avanzada, al mismo tiempo que mejora la velocidad y la eficiencia a través de diseños optimizados y canales de entrenamiento, equilibrando la precisión y el rendimiento \citep{Ultralytics}.

El insumo principal para el entrenamiento del modelo fue la creación de un dataset de 500 imágenes con sus respectivas etiquetas, donde 220 imágenes son referentes a la erosión fluvial y 240 imágenes de un dataset de deslizamientos.  Mediante técnicas de aumento artificial, el dataset final fue de 1204 imágenes. El modelo de YOLOv11 para la detección de erosión fluvial logró una precisión de alrededor del 70\%, a diferencia de los modelos de detección de deslizamientos que tienen una precisión del 80\% al 90\%. Es importante resaltar que al relacionar con la cantidad de imágenes del dataset, se observa una relación directamente proporcional; es decir, entre menos sean las imágenes del dataset, la precisión disminuye como, por ejemplo, en \citep{Zheng}. 

Uno de los principales aportes de este estudio es la estimación del área de la erosión fluvial, así como el despliegue del modelo obtenido en una aplicación web. Este producto permite a usuarios en general la carga de imágenes propias, el ingreso de datos como el tamaño de imagen o píxel, la generación y descarga de las máscaras de erosión fluvial detectadas, y la estimación del área afectada.

\section{EROSCAN} 
\label{sec:eroscan}
Los proyectos de IA generalmente culminan con la etapa de obtener un modelo entrenado y evaluado. A pesar de la complejidad técnica y económica que implica el despliegue de un modelo en producción, uno de los objetivos de este estudio es desarrollar una aplicación web denominada \textit{EROSCAN}, útil para la identificación y estimación de áreas de erosión fluvial y disponible para los usuarios en general.

Para el desarrollo e implementación de la aplicación web se emplearon las siguientes herramientas: el lenguaje de programación Python, el entorno de programación \textit{Visual Studio Code}, el framework de desarrollo web \textit{Streamlit} y la plataforma \textit{GitHub} para la gestión del código implementado.

Streamlit fue seleccionado debido a su facilidad de uso, permitiendo el desarrollo rápido de aplicaciones interactivas sin necesidad de amplios conocimientos. Además, proporciona una interfaz visual intuitiva para desplegar modelos de inteligencia artificial, con funciones integradas como carga de imágenes, generación de gráficos y exportación de resultados.

El proceso de desarrollo de la aplicación se realizó en Visual Studio Code, donde se codificó el script en Python utilizando la librería Streamlit. Posteriormente, el código fue almacenado y organizado en GitHub, lo que facilita el control de cambios y la colaboración.

Para el despliegue en Streamlit Cloud, se enlazó el repositorio de GitHub con esta plataforma en la nube, permitiendo la ejecución automática del código sin necesidad de configuración manual del servidor. Streamlit proporciona un entorno de ejecución en línea que permite visualizar errores y depurar la aplicación en tiempo real.

La Figura \ref{fig:eroscangui} muestra la interfaz gráfica de usuario (GUI) de EROSCAN, donde se pueden apreciar sus principales funcionalidades. La aplicación permite analizar imágenes de manera sencilla y automática. Su ejecución se basa en los siguientes pasos:

\begin{figure}[!htb]
\centering
\includegraphics[width=0.99\textwidth]{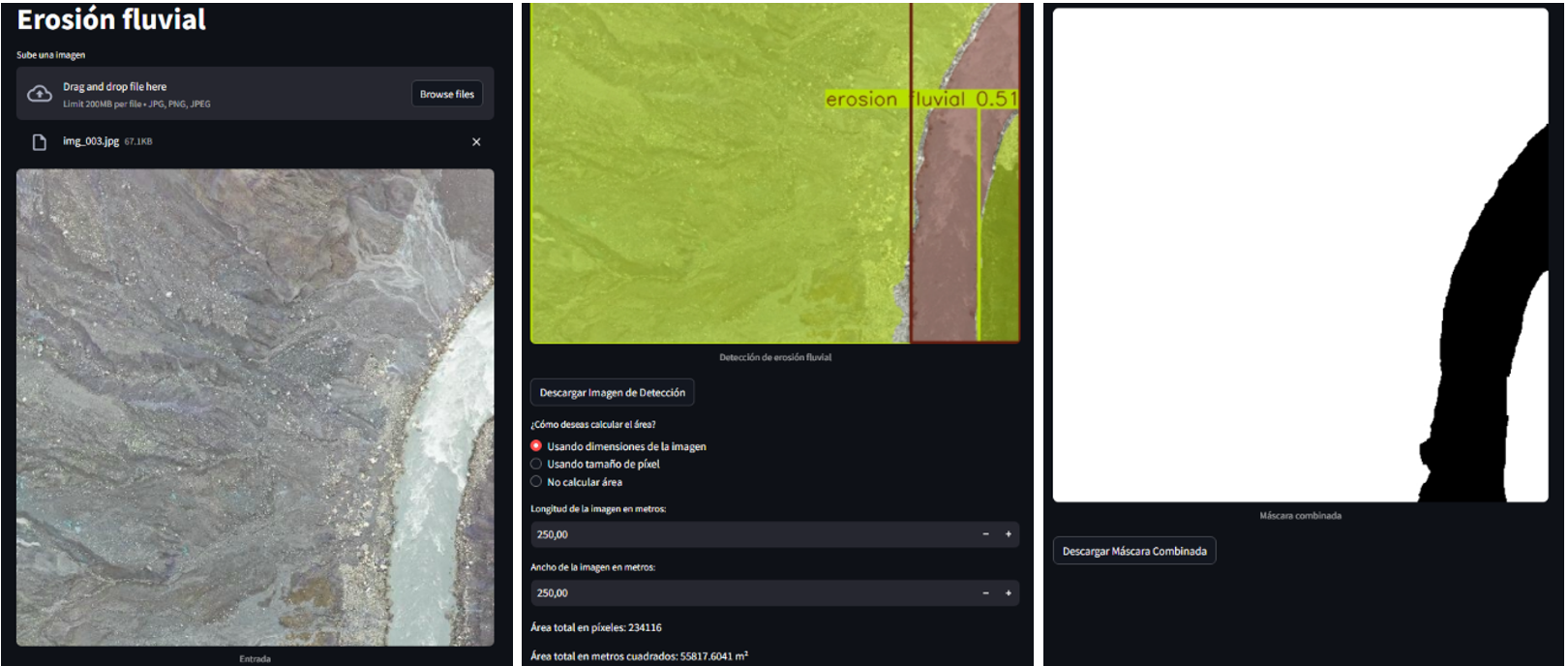}
    \caption{Interfaz de la aplicación web EROSCAN.}
    \label{fig:eroscangui}
\end{figure}

\begin{enumerate}
    \item \textbf{Carga de imágenes:} El usuario sube una imagen en formatos compatibles (.jpg y .png).
    \item \textbf{Procesamiento de la imagen:} Se segmenta automáticamente y se clasifica en categorías relevantes.
    \item \textbf{Cálculo de área:} Se ofrece la opción de calcular el área de la erosión en metros cuadrados (m²) o píxeles (px).
    \item \textbf{Descarga de resultados:} Se generan dos imágenes (.png): una segmentada y una máscara de erosión.
\end{enumerate}

\begin{figure}[!htb]
\centering
\includegraphics[width=0.65\textwidth]{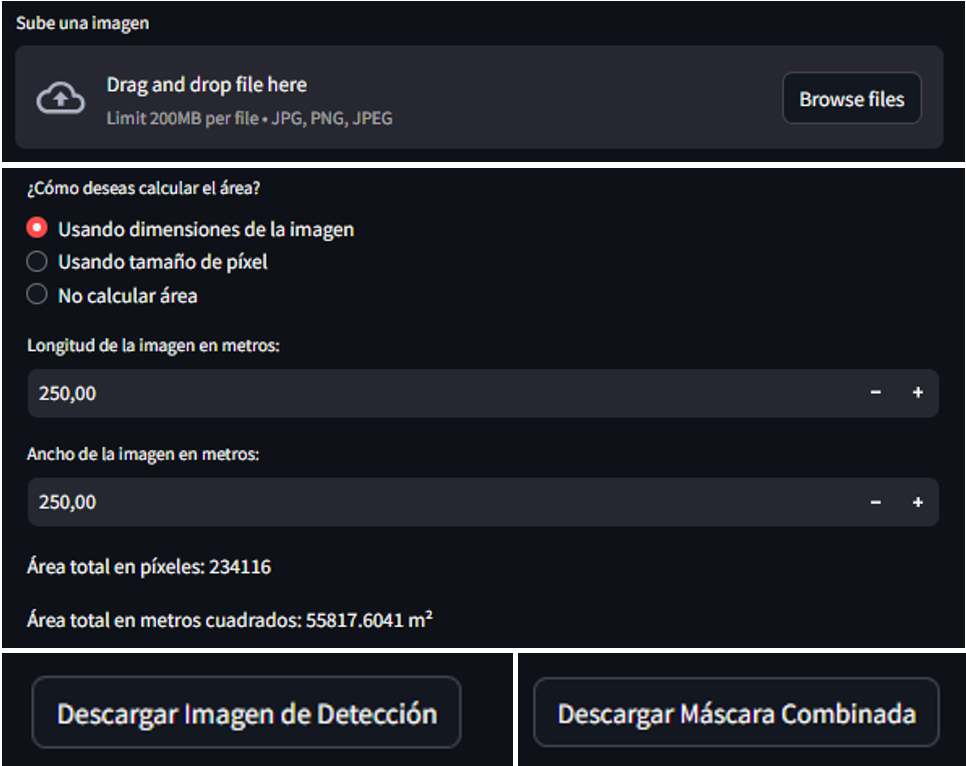}    
    \caption{Botones interactivos y funcionalidades de EROSCAN.}
    \label{fig:eroscanfunction}
\end{figure}

Para mejorar la experiencia del usuario, se implementaron botones interactivos en Streamlit que permiten: a) seleccionar métodos de cálculo de área, y b) descargar las imágenes procesadas, tal como se muestra en la Figura \ref{fig:eroscanfunction}.

La aplicación web no requiere requisitos especiales para su uso. El usuario solo necesita cargar una imagen para que el sistema realice el análisis automáticamente. Se puede acceder a la aplicación desde cualquier dispositivo (computadora o teléfono móvil) a través de un enlace que se proporcionará a los interesados previa solicitud a los autores.

\section{Conclusiones}
\label{sec:conclusion}
Se ha implementado un modelo basado en YOLOv11, en conjunto con la plataforma Roboflow, para la detección y segmentación de zonas de erosión fluvial. Este modelo permite generar máscaras de alta precisión y calcular automáticamente el área afectada. La combinación de técnicas avanzadas de aprendizaje profundo con un conjunto de datos optimizado ha permitido alcanzar un desempeño robusto, adecuado para el análisis de imágenes LiDAR y fotogramétricas.

El desarrollo del dataset específico para la detección de erosión fluvial está compuesto por imágenes representativas cuidadosamente seleccionadas, segmentadas y etiquetadas. Este conjunto de datos ha sido diseñado para optimizar el proceso de entrenamiento del modelo, asegurando una adecuada representación de las variaciones en los patrones de erosión fluvial y permitiendo mejorar la capacidad predictiva del sistema.

Se ha desarrollado una aplicación interactiva utilizando Streamlit, facilitando el acceso a las capacidades del modelo sin necesidad de conocimientos avanzados en procesamiento de imágenes o inteligencia artificial. La plataforma permite la carga de imágenes, la obtención de segmentaciones automáticas y la estimación del área de erosión. Además, integra funcionalidades adicionales como la visualización de los resultados y la descarga de los mismos en formato PNG, lo que amplía su aplicabilidad en estudios de análisis morfológicos.

Para futuros trabajos, es fundamental optimizar la precisión en la detección y segmentación de áreas erosionadas mediante el ajuste de parámetros en modelos preentrenados y la exploración de versiones más recientes de YOLO, como YOLOv12. Se propone ampliar y diversificar el conjunto de datos de entrenamiento con imágenes de mayor resolución y escenarios reales de erosión fluvial, además de incorporar formatos ráster (.tiff) para facilitar la georreferenciación e integración con sistemas de información geográfica. También se contempla el uso de secuencias de video capturadas por drones para extraer fotogramas clave que permitan una estimación más precisa del área afectada.

La combinación de visión por computadora con modelado hidrodinámico permitirá no solo analizar la erosión actual, sino también predecir su evolución considerando factores hidrogeológicos, fortaleciendo la evaluación integral del riesgo y la toma de decisiones en planificación territorial. Este trabajo sienta las bases para el desarrollo de herramientas avanzadas aplicadas al monitoreo geoespacial y la gestión del riesgo, y su implementación futura contribuirá a una detección más precisa y eficiente de la erosión, especialmente en entornos dinámicos y de difícil acceso.

\bibliographystyle{unsrtnat}
\bibliography{references}  







\clearpage
\setlength{\columnsep}{1.5cm}
\twocolumn 
\section*{Autores}

{\flushleft\textbf{Paúl Daniel Maji Dizueta}}

\begin{wrapfigure}{l}{15mm}
    \includegraphics[width=1in,height=1.25in,clip,keepaspectratio]{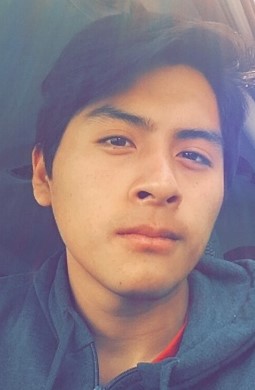}
\end{wrapfigure}\par

Nació en Riobamba, Ecuador, el 21 de septiembre de 1999. Egresado de la carrera de Geología de la Universidad Central del Ecuador. Pasante en el Ministerio de Energía y Minas, en la Subsecretaría de Minería Industrial. Posee experiencia en Geología del Subsuelo y Geología del Petróleo. Participante del Imperial Barret Award (IBA) 2025.\\
\vspace{0.5cm}

{\flushleft\textbf{Marlon Andrés Túquerres Calderón}}

\begin{wrapfigure}{l}{15mm}
    \includegraphics[width=1in,height=1.25in,clip,keepaspectratio]{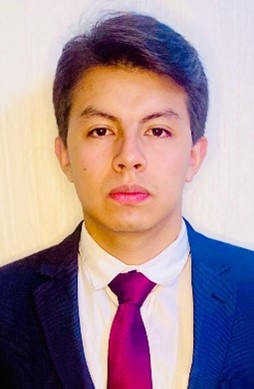}
\end{wrapfigure}\par

Nació en Quito, Ecuador, el 24 de julio de 1998. Egresado de la carrera de Geología de la Universidad Central del Ecuador. Pasante en AZQ – Unidad de Gestión de Riesgos y CELEC EP – Comisión Ejecutora Río Coca. Tiene experiencia en Geotecnia, Geofísica, Mapeo geológico, SIG y Teledetección.\\
\vspace{0.5cm}

{\flushleft\textbf{Marcela Beatriz Valenzuela González}}

\begin{wrapfigure}{l}{15mm}
    \includegraphics[width=1in,height=1.25in,clip,keepaspectratio]{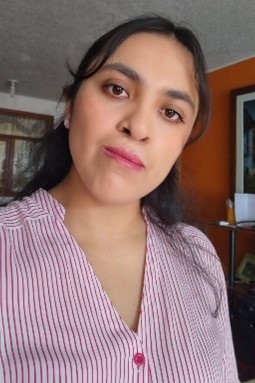}
\end{wrapfigure}\par

Nació en Santo Domingo, Ecuador, el 31 de julio de 1998. Egresada de la carrera de Geología de la Universidad Central del Ecuador. Realizó pasantías en el GAD MIRA, en el área de Gestión de Riesgos y Seguridad Ciudadana. Posee experiencia en mapeo geológico, cartografía geológica y gestión de riesgos.\\

\vfill
\break


{\flushleft\textbf{Stalin Paúl Valencia Molina}}

\begin{wrapfigure}{l}{15mm}
    \includegraphics[width=1in,height=1.25in,clip,keepaspectratio]{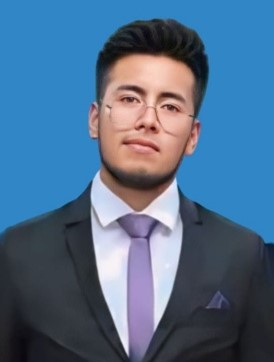}
\end{wrapfigure}\par

Nacido en Quito, Ecuador, el 18 de enero de 2000. Egresado de Geología de la Universidad Central del Ecuador. Experiencia en el sector público: geoanalista, SIG, mapeo y teledetección, así como en la industria privada en empresas petroleras: exploración y desarrollo con registros eléctricos. Líder estudiantil, becario y voluntario en varios proyectos.\\
\vspace{0.5cm}

{\flushleft\textbf{Christian Iván Mejía Escobar}}

\begin{wrapfigure}{l}{15mm}
    \includegraphics[width=1in,height=1.25in,clip,keepaspectratio]{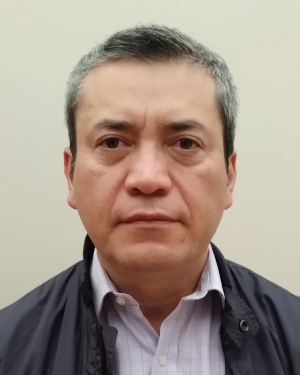}
\end{wrapfigure}\par

Profesor e investigador sobre Inteligencia Artificial en la Universidad Central del Ecuador. PhD en Informática de la Universidad de Alicante, España, 2023. Experiencia en proyectos de Machine Learning y Deep Learning, especialmente en aplicaciones de visión por computadora y procesamiento del lenguaje natural. \\

\end{document}